\documentclass[letterpaper, 10 pt, conference]{ieeeconf}  % Comment this line out if you need a4paper

\IEEEoverridecommandlockouts  
\usepackage{makecell}
\usepackage{array} % For table array
\usepackage{tabularx}
\usepackage{booktabs} % For \toprule etc. in tables
\usepackage{changepage}
\usepackage{newfloat}
\usepackage{tikz}
\usetikzlibrary{shapes.geometric, arrows}
\title{\LARGE \bf
A Soft Robotic Module with Pneumatic Actuation and Enhanced Controllability Using a Shape Memory Alloy Wire
}

\author{Mohammadnavid Golchin$^{1}$
\thanks{$^{1}$Mohammadnavid Golchin is with the Department of Mechanical Engineering and Engineering Science, University of North Carolina,
        Charlotte, NC 28223, USA
        {\tt\small mgolchin@charlotte.edu}}%
}

\begin{document}

\maketitle
\thispagestyle{empty}
\pagestyle{empty}

\begin{abstract}

In this paper, a compressed air-actuated soft robotic module was developed by incorporating a shape-memory alloy (SMA) wire into its structure to achieve the desired bending angle with greater precision. First, a fiber-reinforced bending module with a strain-limiting layer made of polypropylene was fabricated. The SMA wire was then placed in a silicon matrix, which was used as a new strain-limiting layer. A simple closed-loop control algorithm was used to regulate the bending angle of the soft robot within its workspace. A camera was utilized to measure the angular changes in the vertical plane. Different angles, ranging from 0 to 65 degrees, were covered to evaluate the performance of the module and the bending angle control algorithm. The experimental tests demonstrate that using the SMA wire results in more precise control of bending in the vertical plane. In addition, it is possible to bend more with less working pressure. The error range was reduced from an average of $\pm{5}$ degrees to $\pm{2}$ degrees, and the rise time was reduced from an average of 19 seconds to 3 seconds.

\end{abstract}

\section{Introduction}

Soft robotics is a rapidly growing field that is highly regarded by both the academic and engineering communities. Soft robots perform well in terms of environmental adaptation. Similarly to biological organs \cite{c1}, \cite{c2}, they are more flexible. Due to the soft parts of their structure, they can perform tasks such as communicating with the human body, providing clinical and nonclinical rehabilitation, and enhancing safety \cite{c3}. Instead of relying on the relative motion of body components (such as rotor and shaft or piston and cylinder), these robots operate through the deformation of their soft body \cite{c4}. Soft robots are typically actuated by soft actuators, which enable the robot to move and interact with its surrounding environment. These characteristics have provided the possibility of their potential use in medical fields. Intrinsic safety, low weight, relatively easy manufacturing process, and low price are other reasons for the increasing trend of using them in various fields. 
Compressed air and SMA are among the most widely used primary actuators in modular and soft robots with silicon bodies. Compressed air can be considered one of the simplest, most cost-effective, and safest sources for actuating soft robots. This power source is available and supports a good response time. \\
Fiber-reinforced soft pneumatic actuators are among the most successful soft actuators in soft robotics due to their structural strength, range of motion, and power output. Soft actuators, consisting of elastomeric matrices with embedded flexible materials, are also of interest to the robotics community because they are cost-effective and easily customized for specific applications. However, the significant potential of such operators is currently limited because their design is usually empirical \cite{c5}. \\
SMA with unique features such as simple and noiseless activation, lightness, and high power-to-weight ratio can be considered an accurate and appropriate actuation method. Various soft actuators composed of SMA combined with a polymer matrix have been proposed, but their design limitations severely limit their potential for practical applications \cite{c6}–\cite{c9}. SMA wires have a limited linear motion in the range of 4\% to 8\% of their length \cite{c10}, \cite{c11}, but this small change in length can be converted into large deformations out of the plane by embedding them in a polymer matrix \cite{c12}–\cite{c14}. Anisotropic elements can also be added to the polymer matrix to create bending and torsional deformations \cite{c15}, \cite{c16}.\\
Bending is one of the most commonly used forms of motion in robots to access different points in space, rehabilitate damaged joints, and simulate the behavior of various organs. By reviewing past experiences, various geometries that cause bending in different directions can be found. The geometric design with a circular cross-section, presented by Suzumori et al. in 1991, has three degrees of freedom actuated by compressed air. The body of the soft robot was made of fiber-reinforced rubber. Due to the presence of chambers of the same form and volume inside this robot, and by applying air pressure to each separately, the body can bend from the side of the empty chambers. Changing the pressure at the chamber's entrance allows for maneuvering in a three-dimensional space. Despite its lightness and small dimensions, this soft robot has a limited bending range of about 100 degrees with a working pressure of 300 kPa \cite{c17}. Among other geometric designs with a circular cross-section, the 2DOF geometry of the continuum robot developed by Hadi and his colleagues in 2016 is notable. In this robot, the driving forces of the body were provided by SMA springs, which were arranged at a 120-degree angle relative to each other, enabling the robot to maneuver in three-dimensional space. The primary factor causing bending in this continuum robot was the presence of spring holders, which serve as the module body. The robot starts to bend from the same spring by stimulating each spring separately. A standard spring was also used in the center to return the body to its initial position. The bending range of this module was limited to about 45 degrees \cite{c18}. The use of grid geometries with different cross-sections, including rectangular and semi-circular, has also been considered for the ability to create bending in soft actuators. The geometry of the rehabilitation robot developed by Mossadegh and his colleagues in 2014 was in the form of networks, which were bent using compressed air-driven forces through inflation. The final product was bulky. However, with a working pressure of about 200 kilopascals, it can produce a bending angle of 180 degrees \cite{c19}. In 2020, Meng and his colleagues presented a hybrid geometry design using a compressed air actuator and motor-connected wire tendons that produce bending in the soft robot body. In this geometry, one-piece cables were connected to the body from the beginning to the end of the module, and the motor shaft's rotation changed the length. When a bending motion was needed, it was produced by applying compressed air to the silicone body of the module, the amount of which could be controlled by rotating the motor shaft \cite{c20}. In 2019, Lee and his colleagues presented the geometry design of a one-degree-of-freedom bending soft arm by embedding an SMA actuator inside a silicon matrix with a rectangular cross-section. The primary factor causing bending in this geometry was the change in length resulting from the stimulation of the SMA wire, which enabled the wire to slide within the matrix without direct contact with its body. Due to the product's lightness, the 180-degree bending angle was also obtained \cite{c21}. In 2015, Polygerinos and his colleagues at Harvard University examined, analyzed, and compared geometric designs with rectangular, circular, and semi-circular cross-sections. They concluded that the geometric design with a semi-circular cross-section was the most optimal option for creating bending movement. Furthermore, using this geometric design, it is possible to create the maximum amount of bending in the silicone body of the soft robot with pneumatic actuation. This design utilized fibers to strengthen the body and incorporated a strain-limiting layer to facilitate bending motion \cite{c22}.\\
Despite extensive research in soft robotics and its actuators, there are still considerable research problems and opportunities in actuators and geometry designs, selecting appropriate and innovative materials for fabrication, improving deformation and durability, and providing more appropriate bending control capabilities.\\
A soft fiber-reinforced bending robot actuated with compressed air was designed and fabricated \cite{c23}. In this research, to improve the accuracy in the bending control process and overcome problems related to compressed air actuation, a silicon matrix with embedded SMA wire was used as the strain-limiting layer. Angles of 10 to 65 degrees were considered to evaluate the performance of the module and the bending angle control algorithm. Finally, the behavior of both samples represented was analyzed and compared using angle-time graphs.

\begin{figure}[hbt!]
  \centering
  \includegraphics[width=\columnwidth]{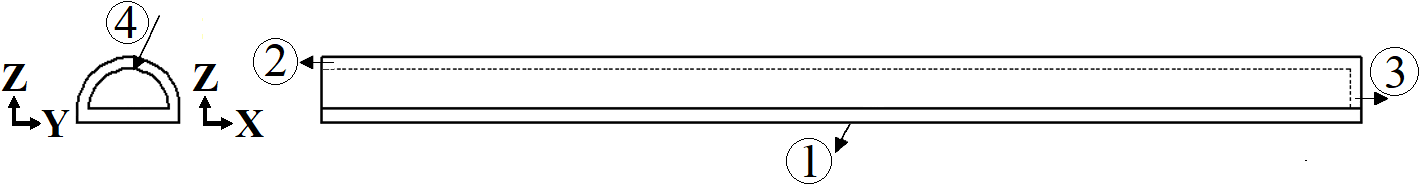}
  \caption{The geometry schematic of the module consists of: 1) the bottom layer, 2) the top layer, 3) the cap, and 4) the air inlet.}
  \label{fig1}
\end{figure}

\section{DESIGN AND FABRICATION METHODS}

\subsection{Design}

Based on the successful geometric design presented, which led to the creation of bending motion in the soft robot \cite{c22}–\cite{c26}, a structure in the form of a semicircular hollow cross section was considered in a rectangular cross section, with the details specified in Figure~\ref{fig1}. Furthermore, the dimensions of the selected body parts are presented in Table~\ref{tab1}.

\begin{table*}[hbt!]
\caption{The soft module dimensions in millimeters.}
\label{tab1}
\centering
\renewcommand{\arraystretch}{1.2}
\begin{tabular}{c c c c c c}
\toprule
\textbf{Length} & \textbf{Width} & \makecell{\textbf{Cap}  \textbf{Thickness}} & \makecell{\textbf{Upper-layer} \textbf{Thickness}} & \makecell{\textbf{Bottom-layer} \textbf{Thickness}} & \makecell{\textbf{Inner} \textbf{Radius}} \\
\midrule
175.1 & 20.5 & 2 & 3.25 & 4.1 & 7 \\
\midrule
\end{tabular}
\end{table*}

Most existing soft robots are made of silicone materials or other types of rubber \cite{c22}, \cite{c27} - \cite{c31}. These materials enable robots to be stretched to high strains, allowing them to expand or contract to different sizes during inflation or deflation. Silicones are often one- or two-part polymers and may contain fillers to improve properties or reduce costs. Unrefined silicone is a gel or liquid that solidifies after processing. The curing method can be vulcanized (using a hardener) or catalyzed. They are generally non-reactive and durable, resisting pressurized environments and temperatures ranging from -55°C to 300°C while maintaining their properties. Due to these properties and ease of fabrication, formability, and flexibility, silicone rubbers can be found in various products, such as industrial applications, cooking, baking, electronics, and medical equipment \cite{c32}. According to the thickness of the model body, two types of silicone (Dragon Skin) with hardness degrees of 20 and 30, as specified in Table~\ref{tab2}, were prepared for use as the robot body.

\begin{table*}[hbt!]
\caption{Silicon technical specifications \cite{c33}.}
\label{tab2}
\centering
\renewcommand{\arraystretch}{1.2} % Optional: adjust row spacing
\begin{tabular}{c c c c c c}
\toprule
\textbf{Type} & \makecell{\textbf{Strain} (\%)} & \makecell{\textbf{Specific Gravity} (kg/m\textsuperscript{3})} &
\makecell{\textbf{Specific Volume}  (m\textsuperscript{3}/kg)} &
\makecell{\textbf{Pot Life} (min)} & \makecell{\textbf{Cure Time}(min)} \\
\midrule
\makecell{Dragon Skin 20} & 620 & 1080 & 25.6 & 25 & 240 \\
\midrule
\makecell{Dragon Skin 30} & 364 & 1080 & 25.7 & 45 & 960 \\
\midrule
\end{tabular}
\end{table*}

By applying air pressure, the internal wall of the module will experience three types of stress: axial, torsional, and radial. The behavior of shape change in this geometry can be seen in two forms: an increase in volume and an increase in length. Fibers are proposed to resist radial stress and prevent body swelling. Kevlar is a type of aramid fiber. This type of fiber is produced in textile industries and is firm, lightweight, and withstands stress, resists corrosion, and heat. Kevlar is used in various applications, including aerospace engineering, automobile brakes, and boat construction. It is usually made in composite form and can be combined with other fibers to produce hybrid composites. The specifications of this fiber are presented in Table~\ref{tab3}.

\begin{table*}[hbt!]
\caption{Kevlar fiber specifications \cite{c34}.}
\label{tab3}
\centering
\renewcommand{\arraystretch}{1.2} % For compact spacing
\begin{tabular}{c c c c c c}
\toprule
\makecell{\boldmath$\sigma$ (MPa)} &
\makecell{\boldmath$\rho$ (kg/m\textsuperscript{3})} &
\makecell{\boldmath$E$ (GPa)} &
\textbf{Poisson Ratio} &
\makecell{\textbf{Strain} (\%)} &
\makecell{\textbf{Thickness} (mm)} \\
\midrule
3600 & 1440 & 60 & 0.36 & 3.6 & 0.75 \\
\midrule
\end{tabular}
\end{table*}

\begin{figure}[hbt!]
  \centering
  \includegraphics[width=\columnwidth]{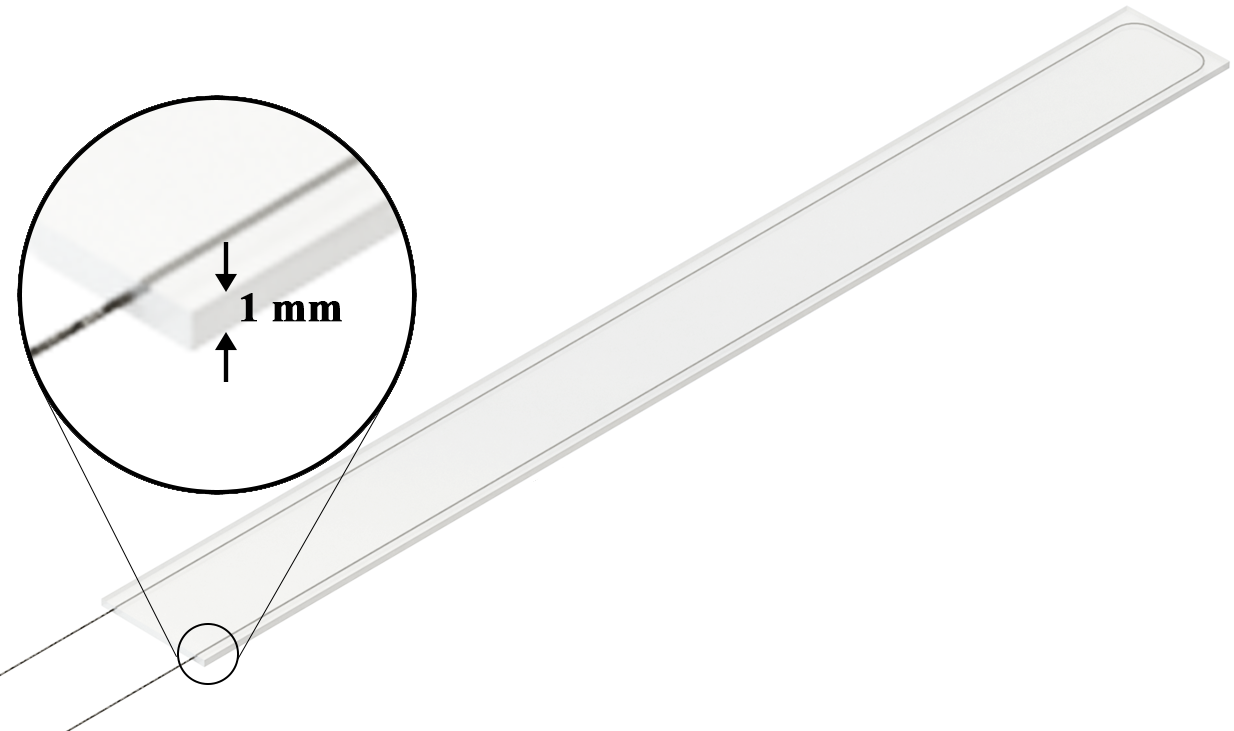}
  \caption{A silicon matrix design that incorporates embedded shape memory alloy wire.}
  \label{fig2}
\end{figure}

Compressed air as an actuator and the twist of Kevlar fiber along the module's body only produce longitudinal motion because of the increased length in the module's body. To create a bending motion in this geometry, it is necessary to prevent stretching of the bottom layer, which has a rectangular cross-section. In this case, the axial stress caused by applying air pressure only results in elongation in the upper layer with a semicircular cross-section. Therefore, by keeping the length of the bottom layer constant, bending occurs in the body to compensate for the elongation of the top layer. For this purpose, nonwoven fabric made of polypropylene is selected and used as a non-stretchable and strain-limiting layer on the lower surface of the body.\\
Enhancing the components of soft robots can increase their capabilities. In this design, smartening of the strain-limiting layer is considered to increase the bending range. Unlike spring types, SMA wires have a limited strain capability. SMA wires in the unactuated state can be stretched by 4\% of their initial length by tolerating a force proportional to their diameter. Among the advantages of using these wires, their dimensions and size can be compared favorably to those of springs. Using the SMA wire inside the soft body minimizes the robot's overall weight and volume change. Controllable deformations can also be achieved in a three-dimensional space and an extensive range of motion \cite{c21}. Therefore, SMA wire can be used as a substitute for the inextensible and strain-limiting layer. The arrangement of the SMA wire, as specified in Table~\ref{tab4}, within a 1 mm silicon matrix with dimensions corresponding to the length and width of the soft robot is shown in Figure~\ref{fig2}.

\begin{table}[bp]
\caption{Shape memory alloy wire specifications.}
\label{tab4}
\centering
\begin{tabular}{c c c c}
\toprule
\textbf{Type} & \makecell{\textbf{Diameter}  \textbf{(mm)}} & \makecell{\textbf{Length}  \textbf{(mm)}} & \textbf{Poisson Ratio} \\
\midrule
Wire & 0.25 & 1000 & 0.33 \\
\midrule
\end{tabular}
\end{table}

\subsection{Fabrication}

To initiate the fabrication process based on the proposed design, various casting methods were employed, including a combination of deposition casting and 3D printing methods. First, the robot geometry was drawn using SolidWorks software, and then the desired mold was designed. To maintain the Kevlar fiber and position it accurately on the body, grooves with the thickness of the fibers were incorporated into the mold. Another mold with module dimensions was designed to create a thin layer to prevent fiber release. Finally, two solid holeless molds were made using 3D printing and polylactic acid filaments. The steps involved in creating the soft bending robot prototype are illustrated in Figure 3. First, a metal rod was inserted into the first mold to maintain the hollow shape of the module. By preparing Dragon 30 silicone, an appropriate amount was poured into the mold, and after the processing time, the first layer of the module slowly separated from the mold. Without removing the metal rod, a 0.85 mm-thick polypropylene strain-limiting layer was placed on the smooth surface of the external wall of the module. Then, a Kevlar fiber was twisted along the body. The module was inserted into the second mold and filled with an appropriate amount of Dragon 20 silicone in the next step. This silicone made the body stiffer and stabilized the fibers on it. After processing time, the module was separated from the second mold. The final step was to create the robot cap and remove the metal rod. 

\begin{figure}[tp]
  \centering
  \includegraphics[width=\columnwidth]{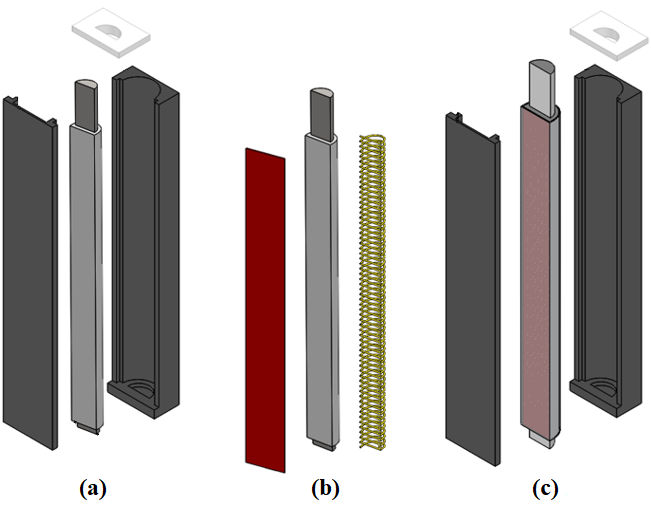}
  \caption{Schematic of the molding steps: (\textbf{a}) represents the first molding process, (\textbf{b}) illustrates the setting of the strain-limiting layer and positioning the Kevlar fibers on the first mold, and (\textbf{c}) depicts the second molding process \cite{c23}.}
  \label{fig3}
\end{figure}

In the fabrication of the developing model, the strain-limiting layer was replaced with a silicone matrix containing an embedded SMA wire to lighten the module and enhance its intelligence. Its location was considered to be the closest to the neutral plane of the module. According to beam theory, the neutral axis is the location where the normal stress is zero. In this case, a higher bending angle is achieved by actuating the SMA wire and producing very low strains. It should be noted that fixing the SMA wire on the body surface was complicated. If it were embedded and stayed in direct contact with silicon, tears would appear in the body because of the low thickness of the layer and the forces resulting from actuation. Therefore, the maximum strain was acquired by applying force to the wire. Then, using fire-resistant glue, a very thin layer was coated on the wire. In addition, the wire was placed in a mold with a depth of 1 mm. The mold was filled with Dragon 30 silicone. After processing time, the fabricated layer was separated from the mold. In this way, the SMA wire is embedded in a thin silicon layer, allowing sliding without direct contact. Similarly to the method of making the first soft robot model, all the steps were repeated, and this silicone matrix replaced the strain-limiting layer made of polypropylene.

\subsection{Experimental Testbed}

\begin{figure}[hbt!]
  \centering
  \includegraphics[width=\columnwidth]{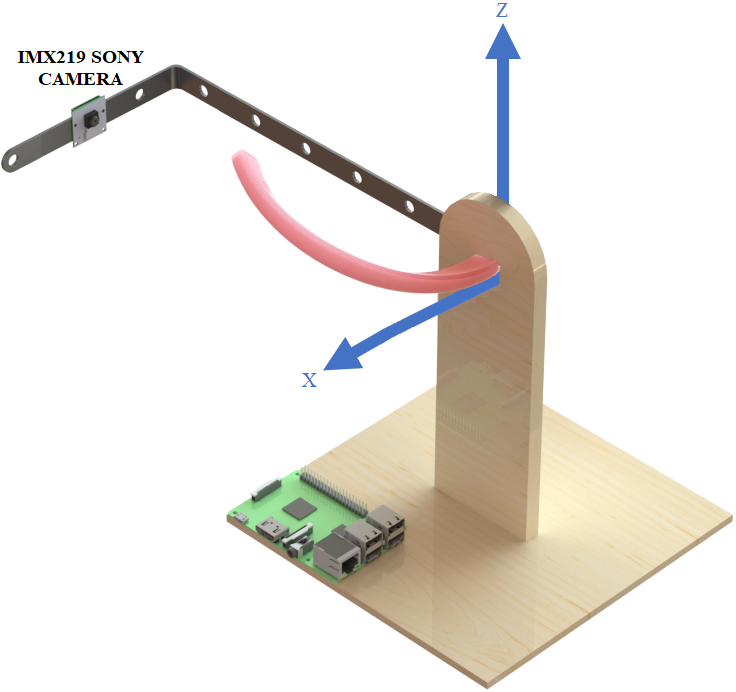}
  \caption{The CAD model for real-time bending angles evaluation system.}
  \label{fig4}
\end{figure}

\begin{figure}[hbt!]
  \centering
  \includegraphics[width=\columnwidth]{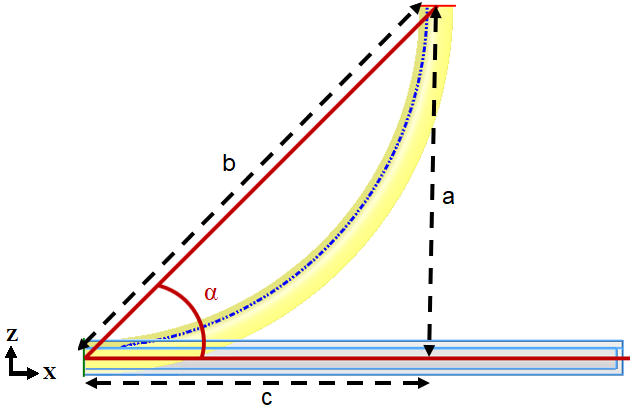}
  \caption{Front view of the module along the Y-axis.}
  \label{fig5}
\end{figure}

To test the fabricated soft modules, it was first necessary to provide their actuation systems. The primary actuator for both robots was a compressed air supply, provided by an air compressor with a capacity of 10 bar. The MPX2200GP air pressure sensor measures the output pressure of the air compressor, with a working pressure range of 0 to 200 kPa. A pneumatic flow control valve, a stepper motor, a driver, and a coupling were used to adjust the amount of input pressure to the system with feedback from the pressure sensor. First, the valve was connected to the motor shaft by a coupling, and then the valve was opened or closed as desired by applying the left or right command to the motor. To measure the degree of bending of the soft robot, a Sony IMX219-160 8MP camera was used to check the module's bending from the front view (in the direction of the Y-axis), as shown in Figure~\ref{fig4}. To measure the bending angle in the plane (XZ) while programming in Python and using libraries and image processing algorithms, first, using a color filter and thresholding technique, the image received from the camera was divided into two separate parts. The color border was then detected around the module, and the pixels outside the border were cleared. In the next step, fixed points were extracted from the corners of the acquired image, connected one after another, and a polygon was formed. Further, the module's image was removed, and only the polygonal image remained, which was updated by changing the robot's shape. Finally, the polygonal image was converted into a triangle, and using the law of cosines expressed in (1), the degree of bending of the soft module was obtained in Figure~\ref{fig5}.

$$
\alpha =cos^{-1}\frac{b^2+c^2-\alpha^2}{2bc}\eqno{(1)}
$$

The Raspberry Pi 3B+ board controlled this process. The function of the bending control algorithm was to provide the desired angle value, within the range of 10 to 75 degrees, as input. The bending angle control algorithm, using the mathematical model proposed by Polygerinos and his colleagues \cite{c22}, calculated the amount of pressure required to reach the desired bending angle and compared it with the output value of the pressure sensor (2).

$$
e_{p}  = P_{pred}  - P_{det} \eqno{(2)}
$$

\begin{figure}[hbt!]
  \centering
  \includegraphics[width=\columnwidth]{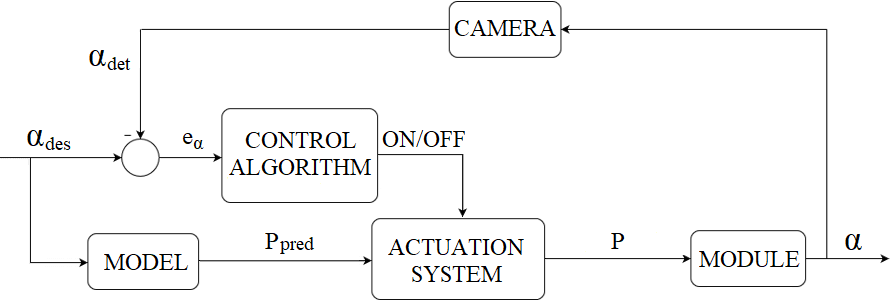}
  \caption{Schematic representation of an experimental closed-loop system for controlling bending angle.}
  \label{fig6}
\end{figure}

Where $e_p$ is the absolute pressure error, $P_{pred}$ is the pressure value obtained through the numerical solution of Polygerinos' mathematical model, and $P_{det}$ is the pressure value read by the air pressure sensor. Following the calculated error, the valve was opened or closed immediately until the desired pressure was achieved. While the opening or closing command was sent to the motor, an increase or decrease in air pressure led to the deformation of the soft robot. In the next step, the images were received in real-time, and the angle in each moment was compared with the desired value. Suppose that the error rate in (3) is positive, the command to open the valve would be sent to the motor, and if the error becomes zero or negative, the command to close the valve would be sent to the motor. Then the amount of applied pressure would be reduced. The schematic of the bending angle control algorithm is shown in Figure~\ref{fig6}.

$$
e_{\alpha} = \alpha_{des}  - \alpha_{det}\eqno{(3)}
$$

\begin{figure}[hbt!]
  \centering
  \includegraphics[width=\columnwidth]{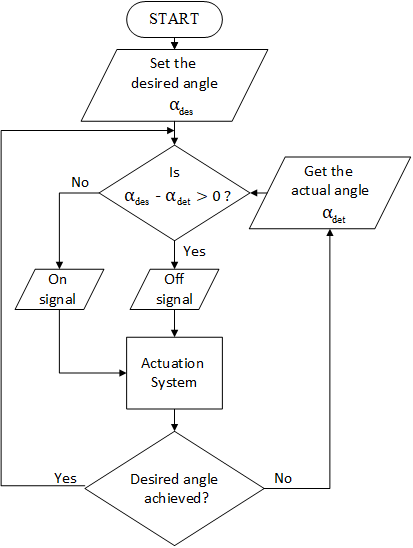}
  \caption{Bending angle control algorithm.}
  \label{fig7}
\end{figure}

In the above equation, $\alpha_{des}$ is the desired angle, $\alpha_{det}$ is the angle read by the camera, and $e_{\alpha}$ is the absolute angular error.
The SMA wire, used as a strain-limiting layer in the soft robot, would be controlled by an L298N driver. When the angular error becomes positive and the command is sent to the motor, the embedded wire will enter actuation mode, causing the pressure valve to open and the robot to bend. In contrast, if the angular error decreases to zero or becomes negative, the actuation of the wire will stop, and the applied pressure will also decrease. The control algorithm is illustrated in Figure~\ref{fig7}.

\begin{figure}[hbt!]
  \centering
  \includegraphics[width=\columnwidth]{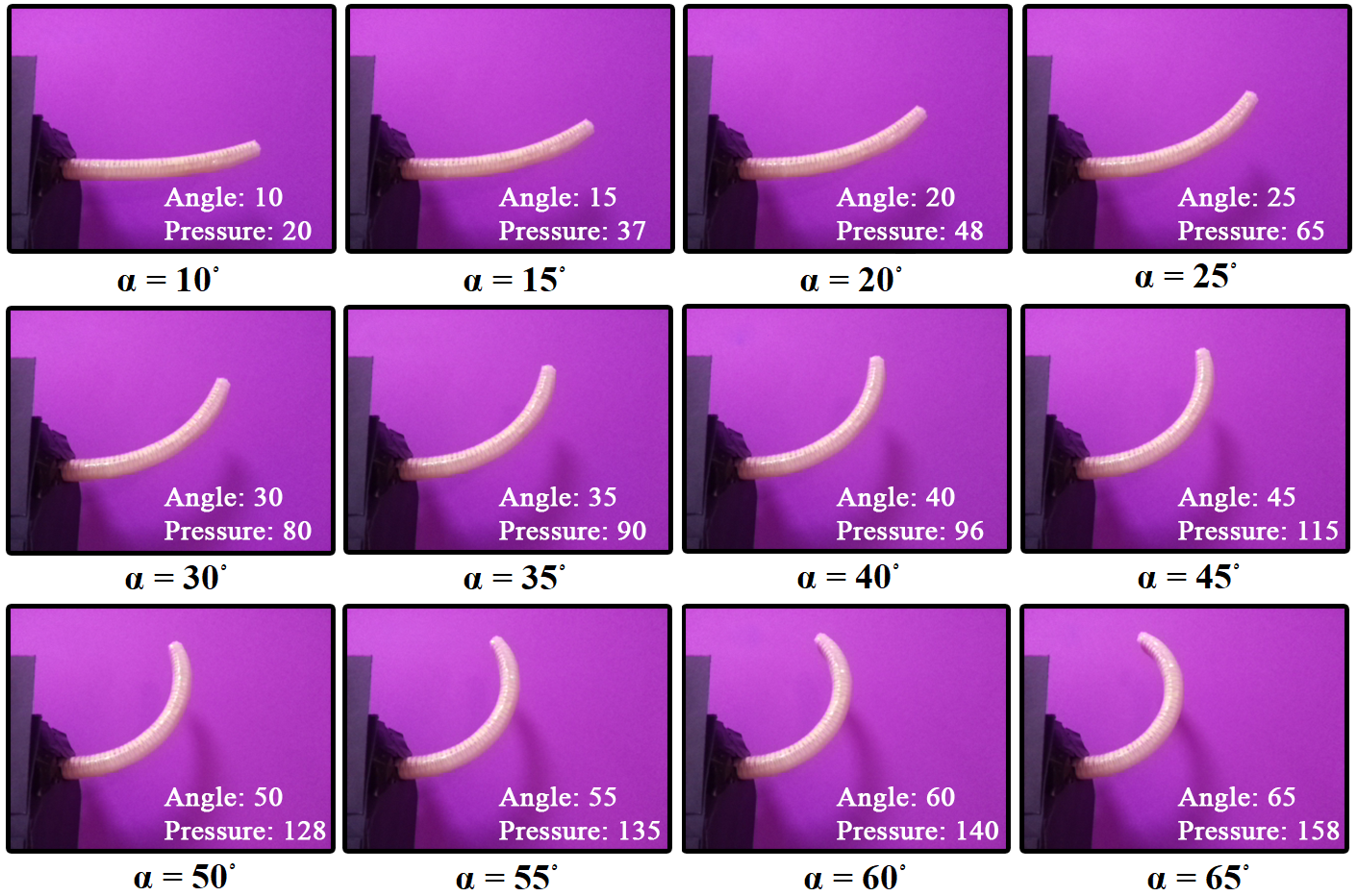}
  \caption{Images captured by the camera showing various orientations of the first soft robotic module actuated by compressed air.}
  \label{fig8}
\end{figure}

\begin{figure}[hbt!]
  \centering
  \includegraphics[width=\columnwidth]{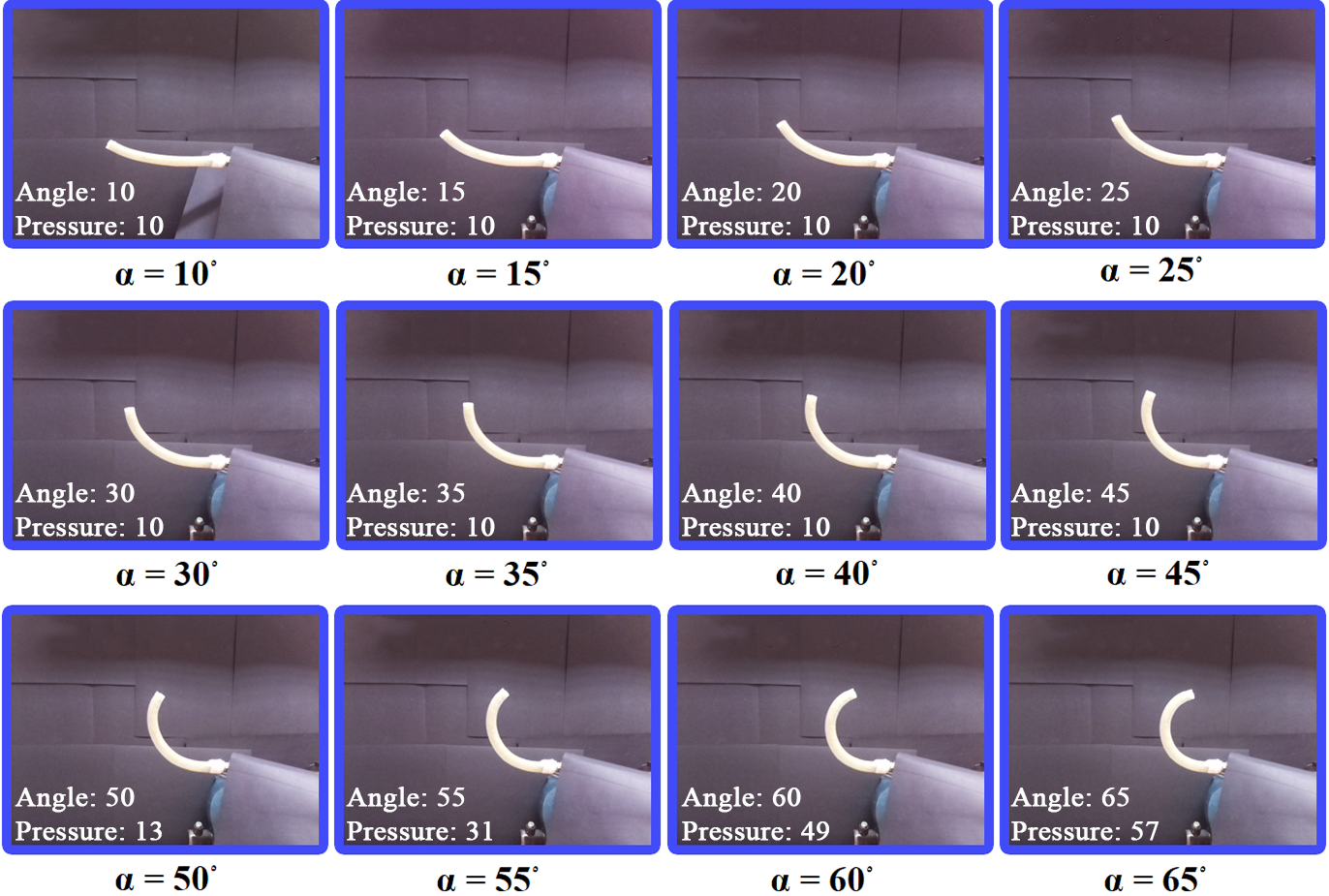}
  \caption{Images captured from the camera show various orientations of the developed prototype of the soft robotic module, which is simultaneously actuated by compressed air and shape memory alloy wire.}
  \label{fig9}
\end{figure}

Experimental tests were conducted to evaluate the performance of the bending angle control algorithm. Both soft robots were expected to bend after applying air pressure and actuating the SMA wire. Figures~\ref {fig8} and~\ref {fig9} show the results related to the feedback of the image received through the camera. Different orientations of both soft robots from angles of 10 to 65 are shown, respectively. As can be seen, the camera detected different angles and both soft robots reached the desired bending angles as expected.

\section{Analysis and Results}

\begin{figure}[hbt!]
  \centering
  \includegraphics[width=\columnwidth]{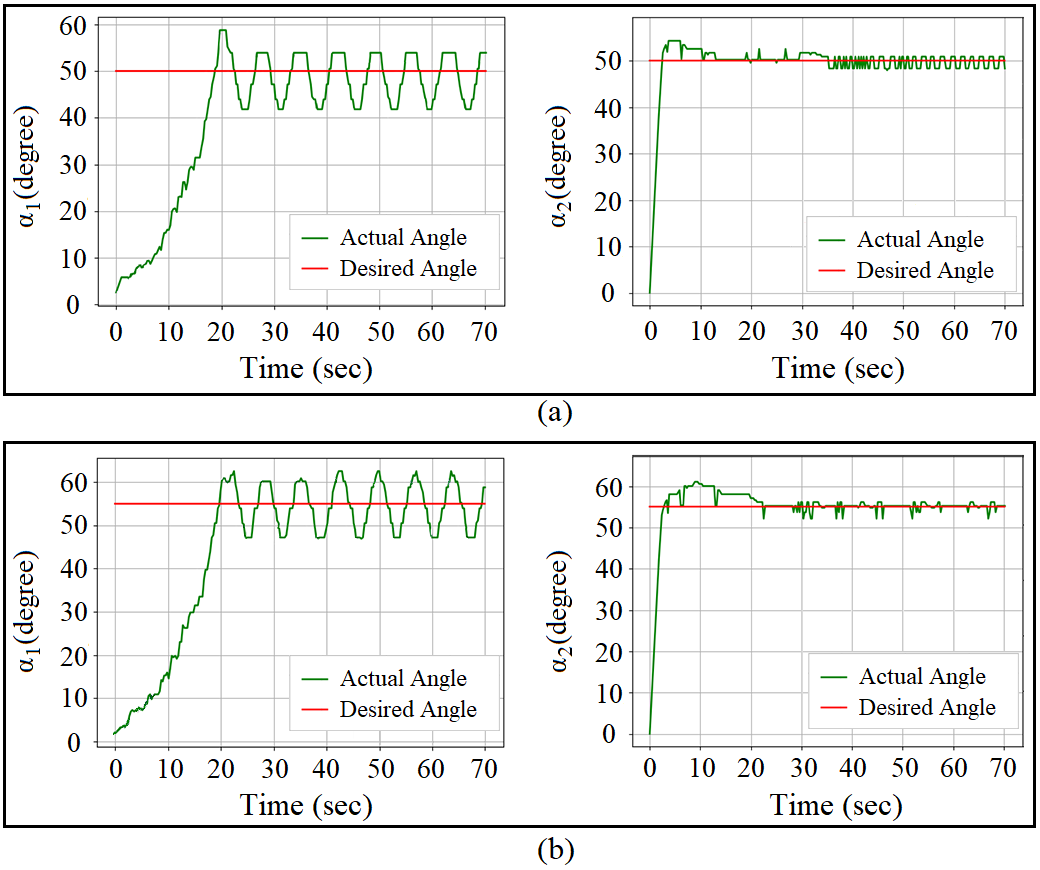}
  \caption{The bending angle evaluation results for both soft robotic modules are presented, with the developed module (shown in the right graphs) being actuated simultaneously by shape memory alloy wire and compressed air. Graph (\textbf{a}) illustrates the results for a desired angle of 50 degrees, while graph (\textbf{b}) shows the results for a desired angle of 55 degrees.}
  \label{fig10}
\end{figure}

\begin{figure}[hbt!]
  \centering
  \includegraphics[width=\columnwidth]{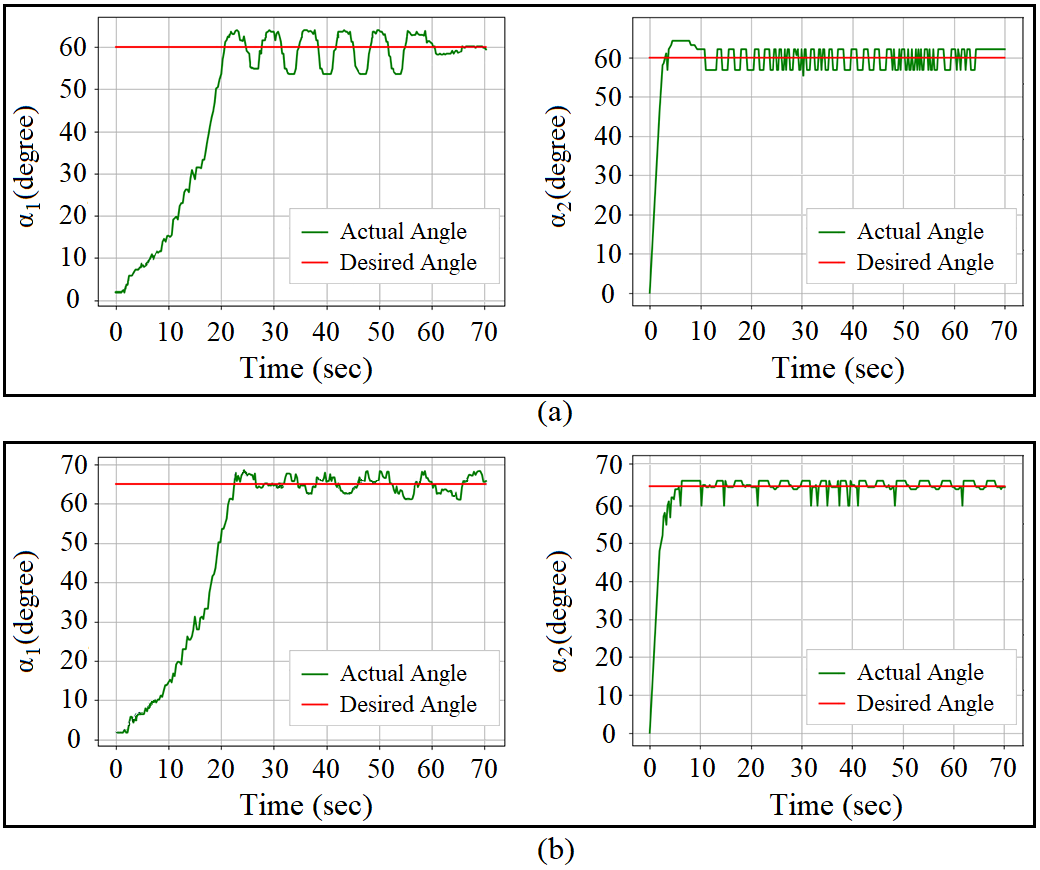}
  \caption{The bending angle evaluation results for both soft robotic modules are presented, with the developed module (shown in the right graphs) being actuated simultaneously by shape memory alloy wire and compressed air. Graph (\textbf{a}) illustrates the results for a desired angle of 60 degrees, while graph (\textbf{b}) shows the results for a desired angle of 65 degrees.}
  \label{fig11}
\end{figure}

The developed module was lighter and more agile, allowing it to achieve the desired angles with less effort. The angle-time graphs of each model at the desired positions of 50, 55, 60, and 65 degrees are presented in Figures~\ref {fig10} and~\ref {fig11}. The duration of each test was 70 seconds. The green line represents the angle detected by the camera, and the red line indicates the desired angle value in each test. In the case where compressed air was used as the only actuator, it can be observed that after approximately 20 seconds, with an average error of ±5 degrees and a delay time of around 14 seconds, the module achieved the desired bending angle.

Meanwhile, the average error would be significantly reduced to ±2 degrees when the SMA wire was used as an actuator in conjunction with compressed air. Additionally, the delay time would be approximately 3 seconds. It should be noted that some of the distortions were caused by processing errors on the part of the camera and the image processing algorithm, as this algorithm was susceptible to ambient light and required calibration after each test. When the SMA wire replaced the strain-limiting layer and was added to the module, the module became lighter and required less air pressure to reach the same positions. In addition, actuating the wire would cause more bending in the module.
According to the results obtained, the developed soft robot is more efficient in controlling the bending angle and power consumption compared to the first module and was able to perform well in following expectations. The intended objectives of this research, including increasing the bending range and improving the controllability of the bending angle, were all realized. The performance of both modules is summarized in Table~\ref{tab5}. \\

\begin{table*}[hbt!]
\caption{Comparison of three modules' performances \label{tab5}}
\centering
\begin{tabularx}{\textwidth}{c c c c c c c c}
\toprule
\textbf{\#} & \textbf{Weight (Kg)} & \textbf{Actuation System}	& \textbf{DOF} & \textbf{Error (°)} & \textbf{Max Pressure (kPa)} & \textbf{Max Angle (°)} & \textbf{Rise Time (s)}\\
\midrule
First Module \cite{c23} & 0.032 & Pneumatic & 1 & ±5 & 210 & 180 & 19\\
\midrule
Second Module \cite{c35} & 0.078 & Pneumatic + SMA Spring & 1 & ±0.85 & 210 & $ 180$ & 12\\
\midrule
Developed Module & 0.026 & Pneumatic + SMA Wire & 1 & ±2 & 210 & $\geq 180$ & 3\\
\bottomrule
\end{tabularx}
\end{table*}

In addition to the weight difference between the first and developed modules, another important consideration is the difference in the maximum available angle. When the maximum working pressure was applied, the first module would bend 180 degrees; however, after some time, the module body would tear and become unusable. In contrast to this behavior, the soft robot developed would achieve this bending angle with a lower working pressure, and no tears would appear on its body.

\section{Conclusion}

In this research, a developed soft robotic module is introduced and presented. In this robot, the combination of compressed air and SMA wire was used to control the bending angle with greater precision. Additionally, an experimental bending angle controller was used to control the bending motions of the module. Then, experimental tests were performed to evaluate the bending angle controller and the effect of the SMA wire on more accurate positioning. The bending angles were identified using real-time image processing. The results showed that despite the complexity of the mechanical behavior, the soft robotic module could maneuver well in the XZ plane and achieve the desired configuration. Further, the developed soft robotic module can maintain accurate positioning with a total error of less than 2 degrees and a rise time of less than 5 seconds. The design of the integration of two compressed air and SMA wire actuators showed that both actuators could minimize system drawbacks while increasing overall capabilities.

\end{document}